\documentclass[11pt]{article}

\usepackage[final]{acl}
\usepackage[normalem]{ulem}
\usepackage{times}

\usepackage{pifont}

\usepackage{latexsym}
\usepackage[T1]{fontenc}
\usepackage[utf8]{inputenc}
\usepackage{microtype}
\usepackage{inconsolata}
\usepackage{graphicx}
\usepackage{verbatim}
\usepackage{amsmath}
\usepackage[dvipsnames]{xcolor}
\usepackage{multirow}
\usepackage{url}
\usepackage{booktabs}
\usepackage{tabularx}
\usepackage{amsfonts}
\usepackage{amssymb}        
\usepackage{nicefrac}
\usepackage{xcolor}
\usepackage{tikz}
\usepackage{xcolor}

\usetikzlibrary{positioning, arrows.meta, calc, fit, shapes.geometric, backgrounds}
\usepackage{pdflscape}      
\usepackage{enumitem}
\usepackage{array} 
\usepackage{stfloats}
\usepackage{cuted}
\usepackage{capt-of}
\usepackage{pifont}
\usepackage{makecell}

\newcommand{\highlight}[2]{\colorbox{#1!20}{\textbf{#2}}}

\title{OpenFinGym: A Verifiable Multi-Task Gym Environment \\for Evaluating Quant Agents}

\newcommand{\cmark}{\ding{51}}                          
\newcommand{\pmark}{\textcolor{gray}{$\boldsymbol{\sim}$}} 
\newcommand{\xmark}{\textcolor{gray}{\textendash}}       

\author{%
  Kaicheng Zhang$^{1,}$\thanks{Equal contribution.} \quad
  Wen Ge$^{2,*}$ \quad
  Lei Jiang$^{3}$ \quad
  Weixin Yang$^{2}$ \\
  \textbf{Jordan Langham-Lopez}$^{3}$ \quad
  \textbf{Jialin Yu}$^{4}$ \\
  \textbf{Lukasz Szpruch}$^{1,}$\thanks{corresponding authors.} \quad
  \textbf{Hao Ni}$^{2,\dagger}$ \\
  $^1$University of Edinburgh \\
  $^2$University College London \\
  $^3$Alan Turing Institute \\
  $^4$University of Oxford \\
  \texttt{K.Zhang-60@sms.ed.ac.uk}  \\
  \texttt{\{wen.ge.25, weixin.yang, h.ni\}@ucl.ac.uk} \\
  \texttt{\{ljiang, jlanghamlopez\}@turing.ac.uk}, \texttt{jialin.yu@eng.ox.ac.uk} \\
  \texttt{L.Szpruch@ed.ac.uk}
}

\begin{document}

\maketitle

\begin{abstract}

Although large language model agents are increasingly applied to quantitative-finance workflows, their evaluation remains fragmented across isolated tasks, while the financial relevance of benchmark tasks is often overlooked. Yet financial workflows are inherently multi-stage, spanning interdependent tasks such as forecasting, strategy construction, risk management, and trading. Existing  platforms typically focus on a single task, and can therefore overstate agent competence and fail to reveal weaknesses in generalization, real-market interaction, and financially meaningful decision-making.
We introduce \textsc{OpenFinGym}, a unified gym environment for quantitative-finance agent development that covers forecasting, market generation, real-time trading, and fraud detection under a single execution and verification interface. \textsc{OpenFinGym} additionally provides an automated task-construction pipeline that turns quantitative finance publications into executable task packages; a containerised runtime with a host-side verifier service that supports scalable agent rollouts and prevents runtime train-test leakage; a paper trading engine with a low-latency data-stream design; deferred-resolution support for long-horizon and event-market forecasts; and integration for SFT and RL post-training. Code is available at \url{https://github.com/DeepIntoStreams/OpenFinGym.git}.

\end{abstract}

\begin{figure*}[t]
  \centering
  \includegraphics[width=\textwidth]{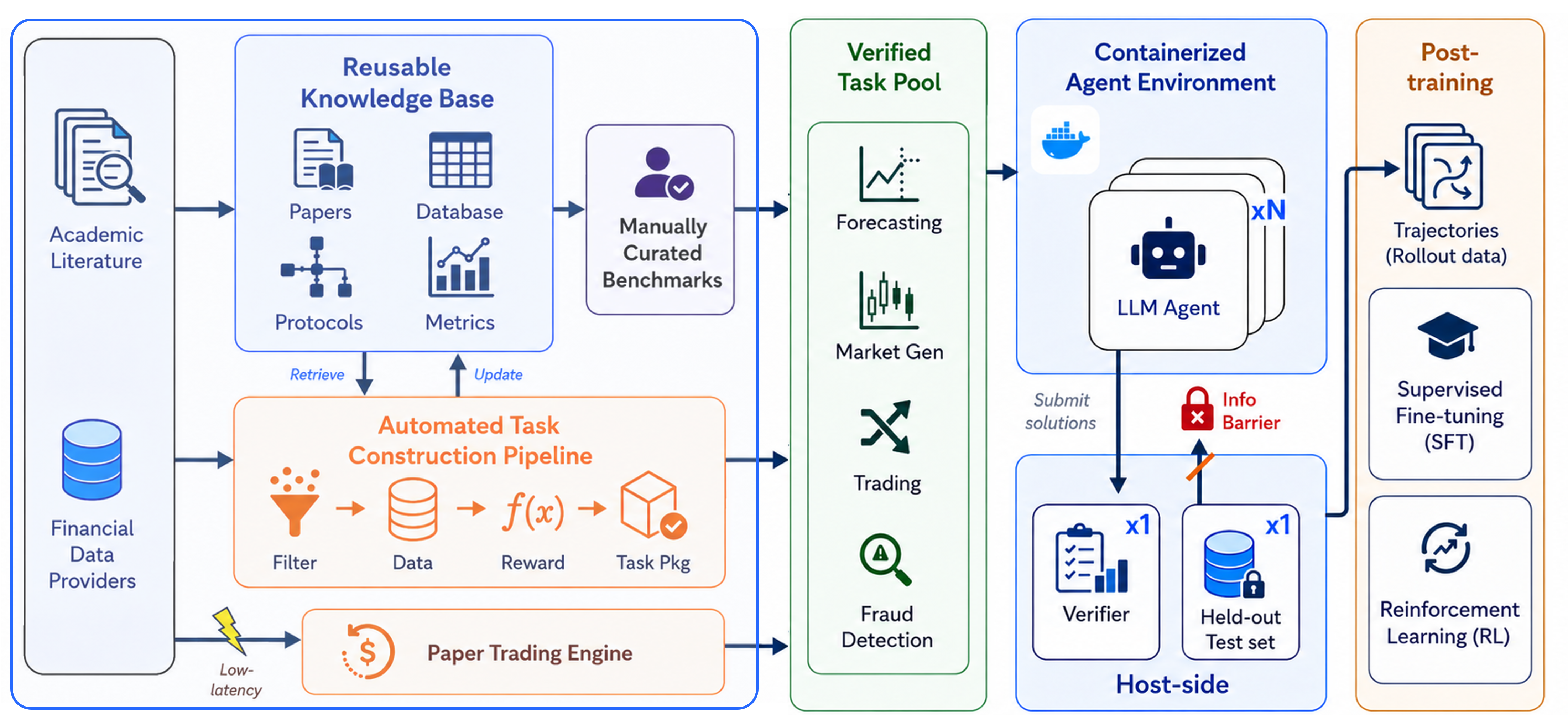}
  \captionof{figure}{A high-level illustration of \textsc{OpenFinGym} architecture}%
  \label{fig:gym_overview}
\end{figure*}

\section{Introduction}
Quantitative finance has, over the past two years, emerged as a testing ground for autonomous large language model agents \citep{zhang2024multimodal,yu2025finmem,yu2024fincon}. Where earlier work investigated whether deep models could outperform classical baselines on a forecasting or trading task \citep{EquityExcessReturnForeca,sun2023trademaster,liu2022finrl}, recent systems treat the LLM itself as a researcher: an agent that selects data, engineers features, trains models, generates predictions or executes trades. This shift changes what evaluation and training infrastructure must support. The task surface should span future price forecasting \citep{jun2024predicting, wang2024ama}, generating synthetic market scenarios for stress testing \citep{hounwanou2025applications}, executing trades against historical and live data \citep{zong2024macrohft, kashif2025lstm}, and detecting anomalies in transaction graphs \citep{li2024sefraud}. It also should span interaction modes, from one-shot batch submission on historical data to step-by-step decision-making on streaming market feeds.

Although some existing platforms such as FinRL-Meta \citep{liu2022finrl}, TradeMaster \citep{sun2023trademaster} supply gym-style trading environments, and domain-general gym environments such as DSGym \citep{nie2026dsgym} and TimeSeriesGym \citep{cai2025timeseriesgym} can handle static tasks such as forecasting, the former are restricted to trading and the latter are not specialised for finance. 
Yet financial tasks are interconnected in practice: forecasts can guide trades, generated scenarios can stress-test strategies, and anomaly signals can inform risk management. No existing platform provides a unified environment for evaluating and training agents across forecasting, generation, trading, and anomaly detection under a common execution and verification interface, which is needed to assess whether agents can perform and generalize across financial workflows. Moreover, in a high-stakes domain such as finance, scalable task construction must be paired with rigorous verification: tasks should reflect meaningful financial workflows while ensuring reproducibility, leakage control, and evaluation against hidden ground truth. 

We introduce \textsc{OpenFinGym}, a unified gym environment for quantitative-finance agent development that closes this gap. \textsc{OpenFinGym} contains a curated suite of 78 tasks across the four task families with an automated, literature-to-task pipeline. Agents run inside containers that exclude held-out test set ground truth, and submit predictions or actions to a host-side verifier service for reward computation. A low-latency data streaming service, a paper trading engine, and a persistent SQLite ledger extend the same interface to real-time trading and long-horizon price and prediction market forecasting. 
We benchmark four frontier LLMs on \textsc{OpenFinGym} and find task-family-specific leaders rather than a single dominant model. We also demonstrate \textsc{OpenFinGym}'s utility as a post-training environment, where SFT and GRPO post training improves the executable generation success rates of Qwen3 base models from 0\% to 100\% on held-out forecasting tasks, and improves the reward by an order of magnitude on certain tasks.

Our contributions are as follows:

\begin{itemize}[topsep=2pt, leftmargin=*, itemsep=1pt]
    \item A curated task suite of 78 verified quantitative-finance tasks spanning four different task families, constructed from prior literature and extended to real-time financial market environment.
    \item An automated task-construction pipeline that converts academic quantitative-finance papers into executable and verifiable task packages.
    \item A containerised runtime that separates agent-visible information from verifier-only ground truth, enabling leakage-resistant evaluation across both batch and sequential tasks.
    \item A gym environment for evaluating and training quantitative-finance agents, which enables supervised fine-tuning and reinforcement learning.
\end{itemize}

\begin{table*}[t]
\centering
\small
\setlength{\tabcolsep}{4pt}
\renewcommand{\arraystretch}{1.2}
\resizebox{\linewidth}{!}{%
\setlength{\tabcolsep}{1mm}
\begin{tabular}{l ccccc cc cc}
\toprule
 & \multicolumn{5}{c}{\textbf{Task coverage}}
 & \multicolumn{2}{c}{\textbf{Runtime}}
 & \multicolumn{2}{c}{\textbf{Agent lifecycle}} \\
\cmidrule(lr){2-6}\cmidrule(lr){7-8}\cmidrule(lr){9-10}
\textbf{System}
 & \makecell{Fore-\\casting}
 & \makecell{Trading\\(offline)}
 & \makecell{Trading\\(real-time)}
 & \makecell{Market\\generation}
 & \makecell{Fraud\\detection}
 & \makecell{Leakage\\control}
 & \makecell{Low-\\latency}
 & \makecell{Auto task\\construction}
 & \makecell{SFT / RL\\integration} \\
\midrule
\multicolumn{10}{l}{\textit{Quantitative finance agent systems}} \\
FinRL-Meta \citep{liu2022finrl}             & \xmark & \cmark & \pmark & \xmark & \xmark & \cmark & \xmark & \xmark & \pmark \\
FinBen \citep{xie2024finben}               & \pmark & \cmark & \xmark & \xmark & \cmark & \xmark & \xmark & \xmark & \xmark \\
INVESTORBENCH \citep{li2025investorbench}         & \xmark & \cmark & \xmark & \xmark & \xmark & \xmark & \xmark & \xmark & \xmark \\
Agent Market Arena \citep{qian2025whenagents}     & \xmark & \xmark & \cmark & \xmark & \xmark & \pmark & \xmark & \xmark & \xmark \\
\midrule
\multicolumn{10}{l}{\textit{General containerised agent gyms}} \\
SWE-Bench \citep{SWE-bench}             & \xmark & \xmark & \xmark & \xmark & \xmark & \cmark & \xmark & \pmark & \xmark \\
TimeSeriesGym \citep{cai2025timeseriesgym}         & \cmark & \xmark & \xmark & \xmark & \xmark & \cmark & \xmark & \pmark & \xmark \\
Harbor \citep{Harbor_Framework}                & \xmark & \xmark & \xmark & \xmark & \xmark & \cmark & \xmark & \xmark & \cmark \\
\midrule
\textbf{\textsc{OpenFinGym} (ours)}
                       & \cmark & \cmark & \cmark & \cmark & \cmark
                       & \cmark & \cmark & \cmark & \cmark \\
\bottomrule
\end{tabular}
}
\caption{Comparison of \textsc{OpenFinGym} against competitive systems across task coverage, runtime properties, and agent-lifecycle support. \cmark{} denotes full support, \pmark{} partial support, and \xmark{} no support.}
\label{tab:system-comparison}
\end{table*}

\section{Related work}

Financial agent environments include reinforcement learning trading platforms such as FinRL-Meta~\citep{liu2022finrl}, TradeMaster~\citep{sun2023trademaster}, quantitative research frameworks such as Qlib~\citep{yang2020qlib}, and market microstructure simulators such as ABIDES-Gym~\citep{amrouni2021abides}. These systems primarily support quantitative research, trading, and simulation rather than executable agentic benchmarking grounded in financial literature. Financial LLM benchmarks, including PIXIU~\cite{xie2023pixiu}, FinBen~\citep{xie2024finben}, FinanceBench~\citep{islam2023financebench} evaluate financial reasoning and analysis abilities on largely static datasets and evaluation protocols. More recent work has focused on executable agent evaluation. SWE-Bench~\citep{SWE-bench}, AgentBench~\citep{AgentBench}, and WebArena~\citep{WebArena} evaluate agents in autonomous environment, while DSGym~\citep{nie2026dsgym}, TimeSeriesGym~\citep{cai2025timeseriesgym}, and Harbor~\citep{Harbor_Framework} extend this paradigm to data-science and time-series with containerised execution settings. \textsc{OpenFinGym} builds on this line of work but targets finance-specific challenges such as temporal leakage and economically grounded evaluation. Concurrent studies further investigate LLM-based financial agents in live or semi-live markets, such as StockBench~\citep{chen2025stockbench}, LiveTradeBench~\citep{yu2025livetradebench}, and When Agents Trade~\citep{qian2025whenagents}. Other works highlight the look-ahead risks, temporal contamination, and benchmark leakage in financial agent evaluation~\citep{benhenda2026lookahead, li2025profitmirage, xu2024benchmarking}. \textsc{OpenFinGym} is positioned at the intersection of executable agent evaluation and quantitative-finance benchmarking through literature-grounded task construction and verifier-isolated evaluation.

\section{Open Financial Gym}
\label{sec:open-financial-gym}

\textsc{OpenFinGym} is designed to support rigorous, scalable, and leakage-resistant evaluation of quantitative-finance agents. 
\textsc{OpenFinGym} covers a broader family of financial workflows, including forecasting, trading, market generation, and fraud detection. 
The benchmark is organised around three components: a reusable quantitative-finance knowledge base, a collection of representative and verified tasks, and a gym-style execution environment for evaluating and training agents. 
The overall architecture is shown in Figure~\ref{fig:gym_overview}.

\subsection{Task Specifications}

\paragraph{Unified task abstraction.}
The quantitative-finance tasks in \textsc{OpenFinGym} are formulated as machine-learning problems. 
Formally, each task is represented by a triplet $(\mathcal{D}, \mathcal{T}, \mathcal{E})$,
where
\begin{itemize}[topsep=2pt, leftmargin=*, itemsep=1pt]
    \item \(\mathcal{D}\) denotes the dataset, including the training and test splits;
    \item \(\mathcal{T}\) specifies the task description, including the data source, input-output interface, prediction or decision objective, required output format, and overall goal of the task;
    \item \(\mathcal{E}\) denotes the evaluation protocol used to assess agent performance. The evaluation protocol may include multiple metrics, together with an aggregation rule that defines the final task score.
\end{itemize}
This formulation provides a unified interface for representing heterogeneous quantitative-finance problems.

Each task is packaged in a standardised format extended from by the Harbor framework~\citep{Harbor_Framework}. 
A task package contains the agent-facing instructions, task interface module, data payload, configuration file used for execution and orchestration, and the evaluation components required by the runtime. 
A key design principle is the separation between information available to the agent and information used only for verification. 
During evaluation, agents may access the training split and the input features of the test split. The corresponding labels, realised outcomes, or hidden evaluation state are withheld and made available only to the verifier. 
This separation reduces leakage risk and allows the same agent to be evaluated across heterogeneous task sources without task-specific glue code.

\begin{table}[t]
\centering
\includegraphics[width=0.99\linewidth]{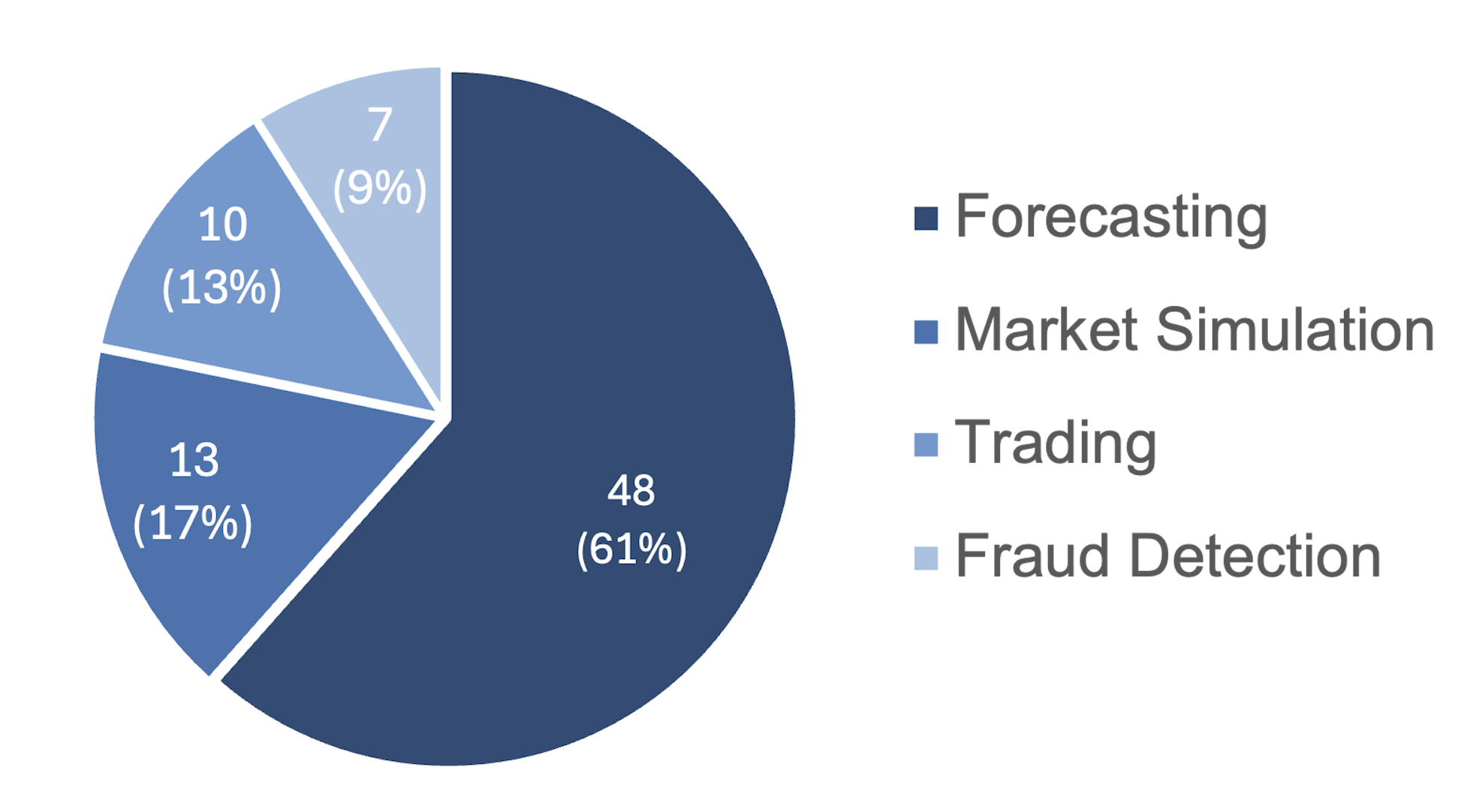}

\vspace{6pt}

\footnotesize
\setlength{\tabcolsep}{3pt}
\renewcommand{\arraystretch}{1.15}
\begin{tabular}{@{} >{\raggedright\arraybackslash}m{0.20\linewidth} >{\raggedright\arraybackslash}m{0.57\linewidth} >{\centering\arraybackslash}m{0.13\linewidth} @{}}
\toprule
\textbf{Task type} & \textbf{Asset Classes} & \textbf{\# Tasks} \\
\midrule
Forecasting & Equities, Commodities, FX, Crypto, LOB, Yield Curve & 48 \\ \midrule
Market Simulation & Equities, FX, Crypto, LOB, Synthetic Processes & 13 \\ \midrule
Trading & Equities, Crypto & 10 \\ \midrule
Fraud Detection & Transactional, Behavioral, and Graph-Structured Signals & 7 \\ 
\bottomrule
\end{tabular}
\caption{Task type distribution (top) and asset class breakdown (bottom) of the curated financial tasks in the gym.}
\label{Tab: task universe}
\end{table}

\paragraph{Task taxonomy.} 
\textsc{OpenFinGym} covers four task families. 
\begin{enumerate}[topsep=2pt, leftmargin=*, itemsep=1pt]
    \item \textbf{Forecasting.}
    Forecasting tasks require an agent to predict future financial quantities, such as returns, prices, volatility, yield-curve movements, or market states, from historical observations and other predictive signals. 
    These tasks are evaluated using statistical metrics such as \(R^2\), RMSE, and directional accuracy, as well as finance-specific metrics that measure the downstream economic value or risk profile of the forecasts.

    \item \textbf{Trading.}
    Trading tasks require an agent to take sequential trading actions on the observed market snapshots. 
    Offline trading tasks evaluate policies on historical data, while real-time trading tasks require sequential interaction with live market observations. 
    Evaluation metrics include profit and loss, Sharpe ratio, drawdown, turnover, and returns with configurable transaction cost and price slippage.

    \item \textbf{Market generation.}
    Market generation tasks evaluate whether an agent can generate realistic financial time series, market states, or scenario distributions. Generated samples are compared with real data using distributional, temporal, and financial diagnostics, including distributional scores, statistic-based scores (e.g., autocorrelation score, marginal distribution score), volatility, tail-risk, and return-dynamics statistics.

    \item \textbf{Fraud detection.}
    Fraud detection tasks are formulated as classification, ranking, or anomaly-detection problems. 
    The objective is to identify fraudulent, suspicious, or anomalous financial activity. Evaluation metrics include AUROC, recall, and threshold-dependent operating metrics.

\end{enumerate}

The task catalogue spans multiple asset classes, including equities, indices, foreign exchange (FX), commodities, crypto-assets, rates, limit-order-book (LOB) microstructure, and prediction markets, as well as different frequencies ranging from intraday to monthly observations and graph-structured data. 
It includes offline tasks derived from the academic literature and real-time tasks in which agents must act before the relevant market or event outcome is observed. 
The resulting task universe is summarised in Table~\ref{Tab: task universe}.

\paragraph{Knowledge base.}
Task construction produces a reusable quantitative-finance knowledge base. 
The knowledge base stores structured information about papers, datasets, task settings, experimental protocols, evaluation metrics, and reusable reward functions. 
This enables datasets and metrics to be reused across related tasks, supports systematic construction of new benchmark instances, and provides the substrate for the automated task construction pipeline described in Section~\ref{sec:methodology}.

\subsection{Agents}

An agent receives the task instructions, the observable data, and the required output contract. 
For batch tasks, the agent typically writes executable code that trains a model on the provided training data and submits predictions for the held-out test inputs. 
For sequential tasks, the agent interacts with the environment through a step-based interface, observes market or portfolio state, and submits actions such as forecasts and trades.

The agent is not required to know how the verifier stores hidden labels or realised outcomes. 
Instead, it interacts with the task through a fixed submission interface. 
This design allows agents with different internal architectures, such as prompted language-model agents, tool-using agents, coding agents, or trained policies, to be evaluated under the same task interface.

\section{Methodology}
\label{sec:methodology}

\textsc{OpenFinGym} has two methodological components. 
The first is a task-construction pipeline that turns quantitative-finance publications and curated benchmark specifications into executable tasks. 
The second is a task-execution framework that runs agents in controlled environments and evaluates their submissions through isolated verifier processes or deferred-resolution services.

\subsection{Task construction}
\label{sec:task-construction}
\begin{figure*}[t]
    \centering
    \includegraphics[width=\textwidth]{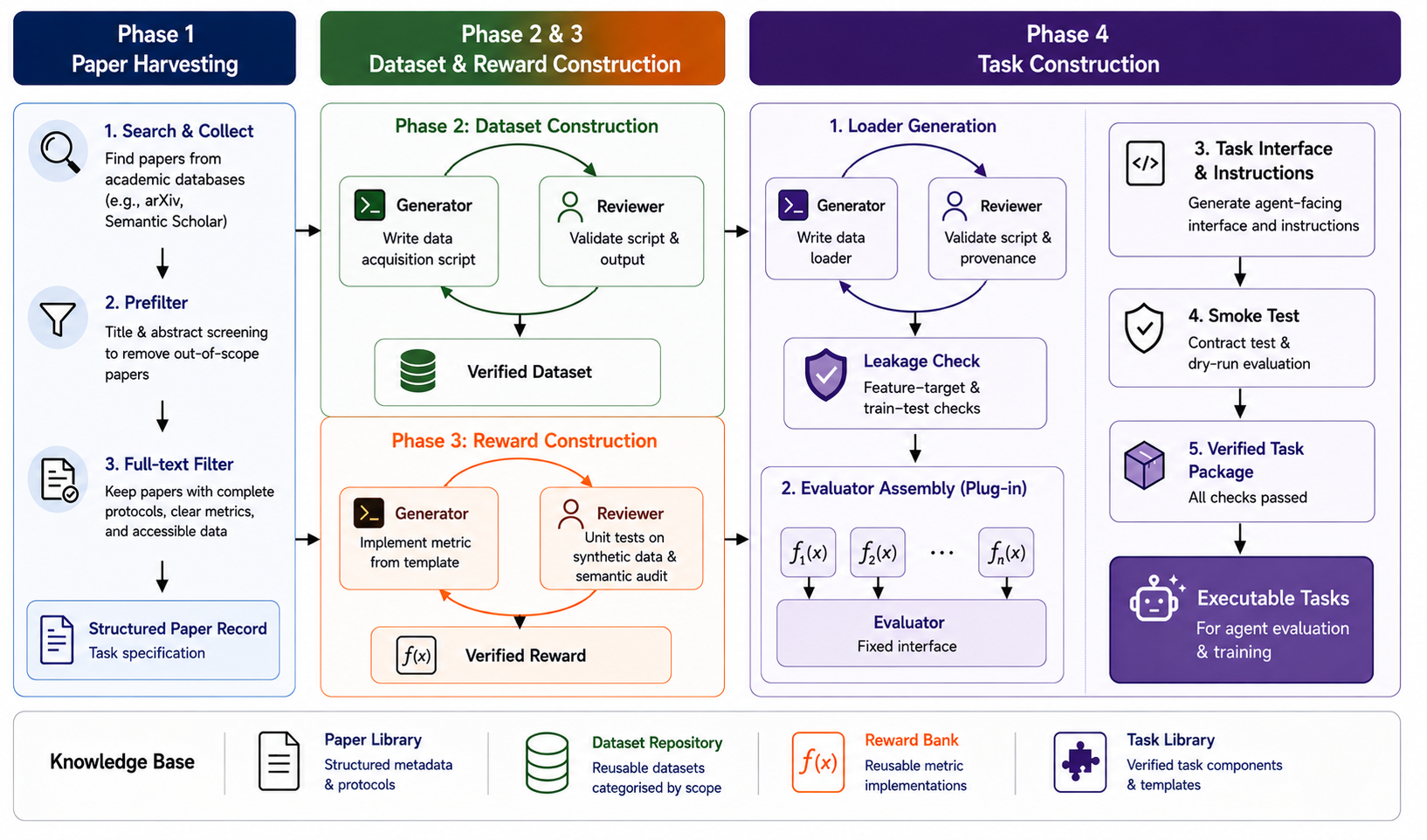}
    \caption{Schematics of the four phases of the task construction pipeline and the knowledge base}
    \label{fig:task_pipe}
\end{figure*}
\textsc{OpenFinGym} constructs tasks through three pathways: automated task generation, curated benchmark implementation, and trading task routing. 
Forecasting and market-generation tasks are well suited to automated construction because they typically follow a common pattern: materialise a dataset, define an input-output split, generate predictions, and evaluate them against held-out targets. 
Trading and real-time tasks require additional infrastructure, including execution engines, market-data providers, portfolio-state tracking, and deferred evaluation. These tasks are therefore implemented through curated harnesses that can be parametrised by paper-specific configurations.

\subsubsection{Automated Task Construction pipeline}
\label{sec:generated-task-pipeline}

\textsc{OpenFinGym} constructs its knowledge base and executable benchmark tasks through a multi-stage pipeline, comprising four main phases: paper harvesting, dataset construction, reward construction, and task generation, as shown in Figure~\ref{fig:task_pipe}.

\paragraph{Paper harvesting.}
The pipeline first collects candidate papers academic finance within a user-specified task scope, then filters them through a two-stage procedure. 
A prefiltering stage screens titles and abstracts to discard out-of-scope papers; 
A full-text filtering stage admits a candidate only if it presents a complete, in-scope experimental protocol together with well-documented evaluation metrics and at least one publicly accessible datasets. 
Accepted papers are summarised into structured records detailing its task setting, datasets, evaluation metrics, and experimental protocol.
These records form the initial substrate of the knowledge base and provide the task specifications for the downstream construction stages.

\paragraph{Dataset and Reward construction.}
For each accepted task specification, the pipeline first check for potential reuses of existing datasets and rewards in the knowledge base. Otherwise, the required artifact is materialised via a generator-reviewer loop, in which an LLM writes the artifact and a reviewer judges it with the aid of unit tests. For dataset construction, the loop runs an LLM-authored download script in a safety sandbox, with the reviewer judging both the script and the materialised payload against the task's metadata. Evaluation metrics undergo similar reuse checks against a reusable reward bank. A metric absent from the bank is generated by the same generator-reviewer loop from a standardised template. It is admitted only after interface checks, unit tests on synthetic tensors and the reviewer's audit. Together these stages populate a reusable quantitative-finance knowledge base of papers, datasets and reward functions organised by task scope.

\paragraph{Task assembly.}
The pipeline assembles the datasets, rewards, and task specifications into a self-contained, \textsc{OpenFinGym}-structured task package. Where applicable, the dataset is partitioned into train and test sets by a task-specific loader generated through the same generator-reviewer loop. The generator step also emits a ground-truth provenance record (e.g. source column, prediction horizon) that the reviewer audits for source-mechanism consistency. A separate feature-leakage test then checks over the loader's output tensors before the loader is accepted. An evaluator is then assembled deterministically from a template by plugging each resolved reward, as specified by the task's metadata. A final generation step writes the agent-facing task interface module, declaring the task's metadata and observation / action spaces, and writes an instruction markdown file detailing task goal, dataset description, and evaluation protocol. The task is installed only after it passes both contract and smoke tests that exercise the full agent--evaluator path. Each package bundles these artifacts with environment configuration files and dataset payload, forming executable benchmark tasks consumable by the \textsc{OpenFinGym} runtime for agent evaluation and downstream training. 

\subsubsection{Generated and curated task packages}
\label{sec:generated-curated-tasks}
The automated pipeline of Section~\ref{sec:generated-task-pipeline} emits a task package from each accepted paper. For forecasting, market-generation and fraud-detection papers, this pipeline materialises the full executable package end to end: dataset, evaluator, task module and instructions. Additionally, we construct offline benchmarks within each scope by manually selecting papers that are widely cited, methodologically reproducible, and accompanied by either public code or a sufficiently detailed protocol. Each task is then implemented and manually verified to reproduce the original literature benchmark as faithfully as possible. The full summary of the curated offline tasks is given in Appendix~\ref{app:curated-tasks}.

For trading tasks, the required execution and market-data infrastructure cannot be reliably built by the automated pipeline alone. The pipeline instead routes the task specification to the curated trading harness and re-exports a task parametrised by the paper-derived configuration (symbols, dates, frequency, prediction horizon, etc.). The harness reused here is the same one used by the manually curated trading tasks, which bundles a configurable execution engine, in-process portfolio-state accounting, and adapters to stock, crypto, and prediction-market data providers.

\begin{table}[t]
\centering
\small
\setlength{\tabcolsep}{4pt}
\caption{Latency and mean refresh rate measurements of the WebSocket (WS) and REST paths of data providers for real-time trading tasks. Both measures are done in a 60\,s window. The WS path incurs an additional $\approx 0.5\,\mu$s buffer lookup time on the agent's side. Measurements were taken on a wirelessly connected laptop from Edinburgh, UK, connected to the Alpaca US-East server and the Binance Tokyo server. Results may vary depend on users' locations and market conditions.}
\label{tab:realtime-latency}
\resizebox{\columnwidth}{!}{
\begin{tabular}{llcccc}
\toprule
\multirow{2}{*}{Provider} & \multirow{2}{*}{Channel} & Latency & Speed-up & Mean refresh & Improvement \\
 & & (ms) & (Latency) & (Hz) & (Freshness) \\
 \midrule
\multirow{2}{*}{Alpaca} & WS & 54.2 & \multirow{2}{*}{$1.90\times$} & 13.53 & \multirow{2}{*}{$9.7\times$} \\
 & REST & 103.1 & & 1.40 & \\
\midrule
\multirow{2}{*}{Binance} & WS & 129.1 & \multirow{2}{*}{$1.86\times$} & 0.67 & \multirow{2}{*}{$40.1\times$} \\
 & REST & 240.5 & & 0.017 & \\
\bottomrule
\end{tabular}}
\end{table}

\subsubsection{Real-time tasks}
\label{sec:realtime-tasks}

Real-time tasks are another case where \textsc{OpenFinGym} relies on a curated harness rather than the automated pipeline. The agent takes trading actions or makes forecasts on streaming market data, and the runtime requirements cannot be reliably built by an automated pipeline alone such as low-latency data streaming, in-process order execution, and long-horizon deferred resolution.

\paragraph{Real-time trading tasks.}
The agent observes live market and portfolio state at each gym step and submits orders through an executor interface. The default is an in-process paper trading simulator that holds positions, cash, and pending orders in memory and fills orders against the latest market snapshot. It supports the common order types and applies configurable price slippage and per-trade transaction costs. The alternative executor routes the same orders to a broker's paper trading sandbox and mirrors the resulting account state locally, keeping account and portfolio reads constant-time on the agent side without a broker round-trip per observation. Further details of the trading implementation are given in Appendix~\ref{apd:paper_trade_engine}.

\paragraph{Latency-aware data stream.}

Each gym step in a real-time trading task reads the current price and order book before the agent acts, and the freshness of those reads to the quality of the agent's actions. A naive REST-per-step approach incurs a network round-trip on every observation and reaches rate limits well below the price refresh rate. \textsc{OpenFinGym} instead serves these data from an in-process buffer that is continuously refreshed by a background WebSocket (WS) subscription. The WS path roughly halves the per-read latency and achieves a peak refresh rate more than an order of magnitude higher than REST, as shown by Table~\ref{tab:realtime-latency}.

\paragraph{Deferred-resolution forecasting tasks.}
\textsc{OpenFinGym} also accommodates price and event market forecasting on real-time data. This raises a different problem: the realised outcome of a long-horizon price forecast or an event market prediction is not available at the moment of submission. The verifier therefore records each submission together with its reference state and the resolution time in a persistent ledger. Once the resolution time has elapsed, a resolver service fetches the realised market price or event outcome from the relevant data provider and computes the configured reward.

\begin{table*}
\centering
\small
\setlength{\tabcolsep}{2mm}
\resizebox{\textwidth}{!}{%
\begin{tabular}{l|cc|cc|cc|cc}
\toprule
\multirow{2}{*}{Model}
& \multicolumn{2}{c|}{Forecasting}
& \multicolumn{2}{c|}{Generation}
& \multicolumn{2}{c|}{Trading}
& \multicolumn{2}{c}{Fraud Detection} \\

& Reward $\downarrow$ & Success (\%) $\uparrow$
& Reward $\downarrow$ & Success (\%) $\uparrow$
& Reward $\uparrow$ & Success (\%) $\uparrow$
& Reward $\uparrow$ & Success (\%) $\uparrow$ \\
\midrule

GPT-4o & 0.052 & 100 & 7.377& 70& \highlight{darkgray}{19.912}& 100& 1.290& 100\\ \midrule
GPT-5.1-codex-mini & \highlight{darkgray}{0.036} & 100 & 11.212&80 & 18.090 & 100 &1.205 & 100\\ \midrule
Haiku 4.5 & 0.058 & 100 & 14.772& 90& 11.490& 100 & 1.147 & 100\\ \midrule
Sonnet 4.6 & 0.053 & 100 & \highlight{darkgray}{7.114} & 90& 12.044 & 100& \highlight{darkgray}{1.630} & 100\\ \bottomrule

\end{tabular}
}
\caption{Agent performance across task families and LLM models. For each task family, we report both the average reward score and the task success rate, aggregated over all tasks within that category. The reward score is computed as a weighted combination of multiple evaluation metrics, providing a comprehensive assessment of overall agent performance. $\downarrow$ represents lower is better while $\uparrow$ represents higher is better.}
\label{tab:agent_results}
\end{table*}

\subsection{Task execution}
\label{sec:task-execution} 

\textsc{OpenFinGym} executes tasks through containerised agent environments and a verifier-based scoring service. Each agent rollout runs inside a container that holds the task instructions, the training data (both inputs and labels), test inputs, task interface, and the Python libraries required for modelling and submission. The held-out test labels are explicitly excluded from the container's bind mounts and are therefore inaccessible to the agent, removing test-set leakage during runtime by construction.

The verifier is a host-side process that loads the held-out test set ground truth into memory at startup and exposes a submission API. The API has a rate limit that prevents the reverse engineering of the held-out ground truth. The verifier process can be shared across every agent container running the same task concurrently, making the setup suitable for large-scale batch evaluation, batch rollouts for RL, and trajectory collection for SFT. On each call, the verifier validates the submission format, computes the reward host-side, and returns the per-metric scores together with a weighted total.

\textsc{OpenFinGym} supports two interaction surfaces: Batch tasks, such as forecasting, use a one-shot submission interface, where the agent submits all predictions in a single call, which are scored immediately or after deferred resolution; Gym-loop tasks, such as trading, use a sequential interface, where the agent repeatedly observes state, takes actions, accumulates a trajectory, and is evaluated at the end of the episode.

\begin{table*}[h]
\centering
\small
\setlength{\tabcolsep}{1.5mm}
\resizebox{\linewidth}{!}{%
\begin{tabular}{l|cc|cc|cc|cc|cc}
\toprule
\multirow{2}{*}{Model}
& \multicolumn{2}{c|}{Commodity}
& \multicolumn{2}{c|}{Crypto}
& \multicolumn{2}{c|}{Equity}
& \multicolumn{2}{c|}{Treasury}
& \multicolumn{2}{c}{FX} \\

& Reward $\downarrow$ & Success $\uparrow$
& Reward $\downarrow$ & Success $\uparrow$
& Reward $\downarrow$ & Success $\uparrow$
& Reward $\downarrow$ & Success $\uparrow$
& Reward $\downarrow$ & Success $\uparrow$ \\
\midrule

Qwen3 1.7b & -- & 0 & -- & 0 & -- & 0 & -- & 0 & -- & 0 \\
\quad SFT  & 0.023 & 60 & 0.539 & 30 & 0.016 & 60 & 1.152 & 60 & \highlight{darkgray}{0.006} & 40 \\
\quad RL   & \highlight{darkgray}{0.020} & 100 & \highlight{darkgray}{0.007} & 100 & \highlight{darkgray}{0.015} & 100 & \highlight{darkgray}{0.099} & 100 & \highlight{darkgray}{0.006} & 100 \\ \midrule
Qwen3 4b   & -- & 0 & -- & 0 & -- & 0 & -- & 0 & -- & 0 \\
\quad SFT  & 0.019 & 70 & 0.007 & 50 & 0.015 & 100 & 1.055 & 60 & \highlight{darkgray}{0.006} & 90 \\
\quad RL   & \highlight{darkgray}{0.018} & 100 & \highlight{darkgray}{0.006} & 100 & \highlight{darkgray}{0.014} & 100 & \highlight{darkgray}{0.090} & 100 & \highlight{darkgray}{0.006} & 100 \\ \midrule
Qwen3 8b   & -- & 0 & -- & 0 & -- & 0 & -- & 0 & -- & 0 \\
\quad SFT  & 0.019 & 90 & 0.007 & 90 & 0.015 & 70 & 0.803 & 70 & 0.006 & 70 \\
\quad RL   & \highlight{darkgray}{0.018} & 100 & \highlight{darkgray}{0.006} & 100 & \highlight{darkgray}{0.014} & 100 & \highlight{darkgray}{0.090} & 100 & \highlight{darkgray}{0.005} & 100 \\ \bottomrule
\end{tabular}
}

\caption{The performance of base model, SFT and RL results on the five held-out forecasting test tasks, averaged over 10 random seeds per task. The first row of each model is the base checkpoint. We report the mean reward ($\downarrow$ lower is better) and the task success rate (\%, $\uparrow$ higher is better). Base success rates are uniformly $0\%$.}

\label{tab:sftrl_results}
\end{table*}

\section{Numerical Results}

\subsection{Agent Performance}

We evaluate a representative subset of OpenFinGym comprising 24 tasks across four families: 5 forecasting tasks, 10 market-generation tasks, 2 trading tasks, and 7 fraud-detection tasks.
Results are summarised in Table~\ref{tab:agent_results}. For each task family, we report both the average reward score and the task success rate. The success rate measures the percentage of tasks for which the agent successfully generates executable submissions that satisfy the task interface and can be evaluated by the OpenFinGym verifier harness.



Several observations emerge from the results. First, all evaluated agents achieve 100\% success rates on forecasting, trading, and fraud-detection tasks, indicating that frontier models can reliably interact with the OpenFinGym interfaces and produce executable quantitative-finance solutions. In forecasting, GPT-5.1-codex-mini performs best with a reward of 0.036, clearly ahead of the other agents whose rewards range from 0.052 to 0.058. In market generation, Sonnet 4.6 gives the strongest result with the lowest reward of 7.114, slightly better than GPT-4o at 7.377 and substantially better than the remaining agents. For trading, GPT-4o achieves the highest reward of 19.912, followed by GPT-5.1-codex-mini at 18.090, indicating stronger trading-strategy quality among the evaluated agents. In fraud detection, Sonnet 4.6 obtains the best reward of 1.630, outperforming GPT-4o by 26.4\%, GPT-5.1-codex-mini by 35.3\%, and Haiku 4.5 by 42.1\%. 

Overall, Table~\ref{tab:agent_results} shows that OpenFinGym reveals task-specific agent strengths rather than a single dominant model: GPT-5.1-codex-mini leads on forecasting, Sonnet~4.6 leads on market generation and fraud detection, and GPT-4o leads on trading. These results demonstrate that OpenFinGym can differentiate agent capabilities across heterogeneous quantitative-finance settings while supporting reliable large-scale automated evaluation.

\subsection{SFT \& RL Results}
\label{sec:sft_rl}

OpenFinGym is integrated with both SFT and RL post-training. To demonstrate its utility as a post-training environment for quant agents, we focus on forecasting tasks under a single-shot generation setting, in which the agent emits its full submission in one pass without iterative tool use.

Our SFT corpus consists of 189 single-shot trajectories collected from Claude Opus~4.7 at low reasoning effort over 21 forecasting training tasks. Agents are encouraged to take classical machine learning approaches over generating neural network architectures and training loops from scratch, as the latter are more difficult for small base models to tackle within a single shot. 

We fine-tune three Qwen3 backbones \citep{yang2025qwen3} of sizes 1.7B, 4B, and 8B with LoRA \citep{hu2022lora} for two to four epochs. The merged LoRA weights then serve as the initialization for RL, where each SFT checkpoint is post-trained with GRPO \citep{shao2024deepseekmath} under SkyRL \citep{cao2025skyrl}. With a batch size of 4 prompts per gradient step and 8 rollouts per prompt, training runs for 50--100 epochs ($\sim$250--500 gradient steps), with reward defined as the negative weighted averages of per-task metrics. Full hyperparameter settings are listed in Appendix~\ref{app:training-integrations}.

Table~\ref{tab:sftrl_results} reports mean reward and success rate on the five held-out forecasting tasks. Base Qwen3 of every size fails to produce an executable submission on any task. SFT closes most of this gap, lifting mean success to roughly $50$--$80\%$. RL then boosts every model to a $100\%$ success rate on every task, with mean reward at or below the corresponding SFT value across the board, including order-of-magnitude reductions on Treasury and on Crypto. We view these results as a demonstration of OpenFinGym's SFT and RL integration.

\section{Conclusion}
We introduced \textsc{OpenFinGym}, a comprehensive benchmark and training environment for quant agents. By combining executable task packages, verification mechanisms, and an automated task-construction pipeline, this gym supports rigorous benchmarking and downstream agent training across diverse financial workflows.

The future works on \textsc{OpenFinGym} includes the extension to a broader range of financial task types, including additional execution settings, risk-management workflows, and decision-making problems beyond the current task families. The existing task-generation pipeline already provides a scalable path for this expansion. Another future direction is to expose \textsc{OpenFinGym}’s authenticated verifier interface as a public evaluation service, allowing external users to evaluate their agents without downloading the full benchmark or accessing test labels. This would enable fair comparison, support leaderboard-style evaluation, and help establish \textsc{OpenFinGym} as a community platform for quant agent developments. 

\section{Limitations}
While \textsc{OpenFinGym} provides a unified environment for evaluating and training quant agents, our current SFT and RL results should be viewed as an initial demonstration rather than a fully optimized training recipe. Due to time and computational constraints, the trajectories used for post-training are limited in both scale and quality, and the backbone agent used in our experiments is relatively lightweight. These choices constrain the attainable gains from supervised fine-tuning and reinforcement learning. Future work will scale trajectory collection, improve trajectory filtering and reward-based data selection, incorporate stronger backbone models, and study larger-scale RL post-training across more diverse financial tasks. 

In this paper, we reduce the risk of data leakage and look-ahead bias by withholding test data and requiring agents to develop and train executable models rather than directly forecast with an LLM. Nevertheless, because some tasks are derived from public papers, the agent may have prior knowledge of effective modelling approaches, making it difficult to disentangle genuine generalization from the influence of such knowledge.

\section{Acknowledgements}
KZ was supported by the EPSRC Centre for Doctoral Training in Mathematical Modelling, Analysis and Computation (MAC-MIGS) funded by the UK Engineering and Physical Sciences Research Council (grant EP/S023291/1), Heriot-Watt University and the University of Edinburgh. WG is supported by the EPSRC through the UCL EPSRC
Landscape Award (UELA) under Grant EP/Z534882/1. LJ, JL and LS acknowledge the support of the UKRI
Prosperity Partnership Scheme (FAIR) under the EPSRC Grant EP/V056883/1 and The Alan Turing Institute. WY and HN are supported by the EPSRC Program Grant [Grant No. UKRI1010] entitled “High
order mathematical and computational infrastructure for streamed data that enhance contemporary
generative and large language models”. HN is also supported in part by the SURE-AI Centre grant 357482, Research Council of Norway. HN and WG are grateful for useful discussions with Tomasz Grzybowski.

\bibliography{custom}

\appendix

\newpage

\section{Paper Trade Engine}\label{apd:paper_trade_engine}
Trading tasks require additional execution and market-data infrastructure that the automated pipeline cannot synthesise from a paper alone. We therefore build a paper trade engine so that the construction of the trading tasks in the task construction pipeline reduces to a configuration change, e.g. asset type, time window, data resolution, etc. The engine is organised around a single executor interface that plugs into the existing financial data providers. The default is an in-process simulator that mirrors portfolio state directly against the latest asset price, as well as applying configurable slippage on market orders and per-trade transaction costs.

Offline trading tasks step over a retrieved historical bar dataset. The engine accepts market, limit, stop, and stop-limit orders under immediate-or-cancel or good-till-cancelled time-in-force, supports batch multi-symbol submissions and explicit order cancellation, and enforces account-level constraints such as cash availability for long entries and a $1\times$ equity cap on short leverage. Orders rejected for constraint violations are surfaced through the next observation step rather than raised as exceptions, letting the agent learn from rejections without crashing the rollout.

Realtime trading tasks act on fresh market data and require a latency-sensitive design. Rather than the naive approach of pulling each observation from a REST endpoint, we subscribe to the data provider's per-trade websocket stream and aggregate the trades into OHLCV bars locally. This decision is motivated by two factors. First, the per-update latency is roughly halved relative to a REST round-trip, and the achievable refresh rate is over an order of magnitude higher in our measurements (Table~\ref{tab:realtime-latency}). Second, the websocket path avoids the tight rate limits of the REST API for frequent data fetch. Each symbol's prices and order book are served from a market-data buffer that a background daemon thread continuously refreshes from the streaming subscription, so an ordinary gym observation step is a constant-time read from memory. A REST refresh is dispatched only as a fallback when the buffer's staleness exceeds a small per-quantity budget; when one is needed, per-symbol calls are fanned out in parallel through a persistent worker pool so that the wall-clock cost is one round-trip rather than one per symbol.

As an alternative to the in-process simulator, and selectable by configuration alone, the engine can route the same orders to a real broker's paper-trading sandbox and mirror the resulting account state locally. The local mirror keeps portfolio queries, e.g. positions, cash, pending orders, and PnL as constant-time lookups on the agent side, and is kept up-to-update by treating the broker's order-event stream as the primary update path with periodic REST calls as gap-fill.

\section{Training Integrations}\label{app:training-integrations}
This appendix lists the hyperparameters and implementation details of the SFT and RL results in Section~\ref{sec:sft_rl}.

\paragraph{Trajectory collection.} The 189 SFT trajectories are produced over 21 training tasks by a one-shot agent with a Claude Opus~4.7 backbone at low reasoning effort. The agent emits a single executable per task, which is executed by the verifier executes and returns an aggregated reward. The system prompt forbids training neural networks from scratch and steers the agent toward ridge, lasso, HistGradientBoosting, LightGBM, ARIMA/VAR, exponential smoothing, and GARCH. Trajectories from the five held-out test tasks are dropped.

\paragraph{Supervised fine-tuning.} Each Qwen3 backbone is adapted with a LoRA module targeting every linear projection: rank $32$ with $\alpha=64$ for the 1.7B model, rank $16$ with $\alpha=32$ for the 4B and 8B models; dropout $0.05$ for 1.7B and 4B, $0.10$ for 8B. Training uses a global batch size of $4$, maximum sequence length $8192$ and AdamW with a warmup-then-cosine schedule. Per-size learning rates are $2{\times}10^{-4}$, $1.5{\times}10^{-4}$, and $1{\times}10^{-4}$; the three models are trained for $184$, $138$, and $92$ optimization steps, corresponding to roughly four, three, and two epochs over the trajectory corpus. This takes $<0.5$ GPUh on a single GH200. As a demonstrative result, we did not tune the training hyperparameter for SFT. The trained LoRA weights are then merged into the full-parameter checkpoint that initializes RL.

\paragraph{Reinforcement learning.} GRPO runs under SkyRL and vllm with full-parameter updates. Rollouts use temperature $1.0$ and top-$p$ $1.0$, with sequence length capped at $8192$ tokens. We apply a KL penalty of $0.01$ against a frozen reference policy together with token-level truncated importance sampling clipped at $2.0$. Training takes roughly $1-4$ GPUh depending on the model size. The RL system prompt is identical to the one used at trajectory collection and SFT.

\paragraph{Reward transform.} The verifier's per-task scalar is a lower-is-better weighted aggregate of task-specific metrics. We remap it to a maximization-friendly RL reward by negating and clipping: scoreable trials receive $r = -\min(\mathrm{loss},\, 5)$, while failed or unparseable submissions receive a fixed $r = -6$ so that any scoreable trial is strictly preferred over any failure.

\section{Additional Experiments}
Since the benchmark uses heterogeneous metrics with different scales and preferred directions, the full-corpus comparison uses within-task relative ranks (lower is better). We evaluate every curated non-real-time task; real-time tasks are excluded because their scores depend on contemporaneous market conditions.

\paragraph{Expanded task coverage.}
Table~\ref{tab:rebuttal-all-ranks} reports average relative ranks across all curated non-real-time tasks. GPT-5.1-codex-mini has the best overall rank, narrowly ahead of Sonnet~4.6, while the family-level results show that no model dominates all four families.

\begin{table}[t]
\centering
\small
\resizebox{\columnwidth}{!}{%
\begin{tabular}{lccccc}
\toprule
Model & Fcast. & Gen. & Trade & Fraud & Avg. \\
\midrule
GPT-5.1-codex-mini & 1.688 & 2.400 & 2.000 & 2.857 & \textbf{2.236} \\
Sonnet 4.6 & 2.667 & 2.100 & 2.625 & 1.571 & 2.241 \\
GPT-4o & 2.292 & 2.800 & 2.500 & 2.857 & 2.612 \\
Haiku 4.5 & 3.333 & 2.400 & 2.750 & 2.714 & 2.799 \\
\bottomrule
\end{tabular}}
\caption{Average within-task model ranks over all curated non-real-time tasks. Lower is better.}
\label{tab:rebuttal-all-ranks}
\end{table}

Table~\ref{tab:rebuttal-significance} reports two-sided pairwise sign tests and Wilcoxon signed-rank tests on paired task-level outcomes pooled across the four families. GPT-5.1-codex-mini significantly outperforms GPT-4o and Haiku under both tests. GPT-4o outperforms Haiku under the sign test only; the remaining comparisons are not significant.

\begin{table}[h]
\centering
\small
\resizebox{\columnwidth}{!}{%
\begin{tabular}{lccc}
\toprule
Pair & Sign $p$ & Wilcox. $p$ & Sig. \\
\midrule
GPT-5.1--Haiku & $2.7\times10^{-5}$ & $8.7\times10^{-3}$ & Both \\
GPT-4o--Haiku & $7.9\times10^{-5}$ & $0.18$ & Sign only \\
GPT-5.1--GPT-4o & $6.3\times10^{-3}$ & $4.0\times10^{-2}$ & Both \\
GPT-5.1--Sonnet & $0.14$ & $0.37$ & No \\
Sonnet--Haiku & $0.16$ & $0.18$ & No \\
GPT-4o--Sonnet & $0.21$ & $0.79$ & No \\
\bottomrule
\end{tabular}}
\caption{Pairwise significance tests on paired task-level performance.}
\label{tab:rebuttal-significance}
\end{table}

\paragraph{Repeated-run results.}
Table~\ref{tab:rebuttal-family-examples} gives one representative task per family as mean $\pm$ standard deviation over three runs. The results also expose sensitivity to reward aggregation: the best model on one component metric need not be best on another. Per-task reward weights are configurable and documented in \texttt{curated\_task\_paper\_matrix.md}.

\begin{table*}[t]
\centering
\small
\resizebox{\textwidth}{!}{%
\begin{tabular}{lcccc}
\toprule
Model & Forecasting $\downarrow$ ($\times10^{-2}$) & Generation $\downarrow$ & Trading $\uparrow$ & Fraud $\uparrow$ \\
\midrule
GPT-4o & $1.93\pm0.0086$ & $1.74\pm0.022$ & $37.8\pm7.7$ & $0.655\pm0.27$ \\
GPT-5.1-codex-mini & $1.89\pm0.13$ & $1.52\pm0.18$ & $30.8\pm0.0$ & $0.928\pm0.0098$ \\
Haiku 4.5 & $2.00\pm0.00075$ & $1.46\pm0.077$ & $26.3\pm8.2$ & $0.423\pm0.29$ \\
Sonnet 4.6 & $1.88\pm0.069$ & $1.41\pm0.0031$ & $16.7\pm16.0$ & $0.898\pm0.012$ \\
\bottomrule
\end{tabular}}
\caption{Representative results from each task family. Arrows indicate the preferred direction.}
\label{tab:rebuttal-family-examples}
\end{table*}

\paragraph{SFT and RL robustness.}
Base-model failures most often arise from hallucinated packages or function arguments and from output-shape mismatches. Table~\ref{tab:rebuttal-sft-rl} reports mean $\pm$ standard deviation over ten seeds on the five held-out forecasting tasks. SFT removes the observed execution failures, and RL further improves numerical reward. Pooling the paired SFT and RL means over three model sizes and five tasks yields a Wilcoxon signed-rank test with $p=0.0007$.

\begin{table*}[t]
\centering
\small
\resizebox{\textwidth}{!}{%
\begin{tabular}{lccccc}
\toprule
Model & Commodity & Crypto & Equity & Treasury & FX \\
\midrule
Qwen3-1.7B SFT & $0.0234\pm0.0076$ & $0.5393\pm0.9214$ & $0.0158\pm0.0021$ & $1.1519\pm0.9723$ & $0.0058\pm0.0000$ \\
Qwen3-1.7B RL & $0.0198\pm0.0009$ & $0.0073\pm0.0005$ & $0.0149\pm0.0004$ & $0.0994\pm0.0150$ & $0.0057\pm0.0001$ \\
Qwen3-4B SFT & $0.0193\pm0.0012$ & $0.0074\pm0.0016$ & $0.0151\pm0.0001$ & $1.0548\pm0.4748$ & $0.0057\pm0.0002$ \\
Qwen3-4B RL & $0.0179\pm0.0002$ & $0.0064\pm0.0004$ & $0.0138\pm0.0000$ & $0.0898\pm0.0000$ & $0.0056\pm0.0001$ \\
Qwen3-8B SFT & $0.0194\pm0.0008$ & $0.0070\pm0.0010$ & $0.0148\pm0.0003$ & $0.8035\pm1.0008$ & $0.0058\pm0.0001$ \\
Qwen3-8B RL & $0.0178\pm0.0001$ & $0.0062\pm0.0000$ & $0.0139\pm0.0002$ & $0.0898\pm0.0000$ & $0.0054\pm0.0001$ \\
\bottomrule
\end{tabular}}
\caption{Held-out forecasting reward over ten seeds (mean $\pm$ standard deviation); lower is better.}
\label{tab:rebuttal-sft-rl}
\end{table*}

\section{Curated Tasks Catalogue}\label{app:curated-tasks}

This appendix catalogues the manually curated and verified forecasting, 
generation, and fraud detection tasks shipped with OpenFinFym. Each task fixes deterministic train/validation/test splits, a supervised or sample contract that pins the agent's output shape, and a verifier-enforced evaluator whose metric definitions are ported from the source paper.

\paragraph{Dataset provenance.} Every curated task is either reproduced from a publicly released source-paper benchmark~\cite{MOSAIC,TSAgent,HighRank,TimeGAN} following their Apache licenses, or rebuilt from open public data feeds---Yahoo Finance~\cite{YahooFinance}, EIOPA~\cite{EIOPA}, and U.S.\ Treasury par yields~\cite{TSFM,EquityExcessReturnForeca,Nifty50MultiFactor,GobalIndexVolForeca,PCFGAN,YieldCurveForeca} in a way Consistent with their intended uses; no proprietary or paywalled feed is required at any stage. Synthetic tasks (Ornstein–Uhlenbeck (OU) and rough-volatility data ~\cite{PCFGAN}, Tail-GAN~\cite{TailGAN}, and the multivariate fractional Brownian Motion (fBM) dataset~\cite{HighRank}) regenerate their data based on the model parameters specified in the corresponding source papers. Fraud-detection tasks reuse established node-anomaly benchmarks~\cite{zheng2025cluster} together with a synthetic graph benchmark~\cite{li2024sefraud}. Per-task source citations appear in the Source column of Tables~\ref{tab:curated-forecasting},~\ref{tab:curated-generation}, and~\ref{tab:fraud_detect}.

\subsection{Metric definitions and references}\label{app:curated-metrics}

Table~\ref{tab:appendix-metrics} defines the metrics abbreviations used throughout this appendix. Unless otherwise stated, scoring metrics are losses for which smaller values are better.

\begin{table}[h]
\centering
\small
\begin{tabularx}{\linewidth}{l X}
\toprule
\textbf{Symbol} & \textbf{Definition} \\
\midrule
\multicolumn{2}{l}{\emph{Point-forecast errors}} \\
MSE & mean squared error. \\
RMSE & root mean squared error. \\
MAE & mean absolute error. \\
MAFE & mean absolute forecast error (per-horizon MAE used in the volatility-graph task). \\
MAPE & mean absolute percentage error. \\
SMAPE & symmetric MAPE. \\
$R^{2}$ & coefficient of determination. \\
$R^{2}_{\rm OOS}$, $R^{2}_{\rm zero}$ & out-of-sample $R^{2}$ / $R^{2}$ vs.\ a zero forecast. \\
\midrule
\multicolumn{2}{l}{\emph{Statistical tests}} \\
DM & Diebold-Mariano forecast-accuracy test~\citep{DieboldMariano1995, EquityExcessReturnForeca}. \\
NW $t$-stat & Newey-West HAC $t$-statistic~\citep{NeweyWest1987, TSFM}. \\
\midrule
\multicolumn{2}{l}{\emph{Financial-risk diagnostics}} \\
Sharpe & annualised Sharpe ratio. \\
VaR$_{\alpha}$, ES$_{\alpha}$ & Value-at-Risk / Expected Shortfall at level $\alpha$. \\
$\Delta$Sharpe & asset-mean $|\text{Sharpe}_{\rm pred}-\text{Sharpe}_{\rm true}|$. \\
$\Delta$VaR & asset-mean $|\text{VaR}_{\rm pred}-\text{VaR}_{\rm true}|$. \\
$\Delta$ES & asset-mean $|\text{ES}_{\rm pred}-\text{ES}_{\rm true}|$. \\
Kupiec test & unconditional coverage test for VaR~\citep{Kupiec1995, TailGAN}. \\
\midrule
\multicolumn{2}{l}{\emph{Interval forecasting}} \\
PICP & prediction-interval coverage probability (target $\approx 0.95$)~\citep{Khosravi2010, YieldCurveForeca}. \\
MPIW & mean prediction-interval width (tighter is better, conditional on PICP)~\citep{Khosravi2010, YieldCurveForeca}. \\
\midrule
\multicolumn{2}{l}{\emph{Generative-model evaluators}} \\
Disc.\ score & post-hoc discriminative score: train / test AUC of a classifier separating real from generated paths~\citep{TimeGAN, PCFGAN}. \\
Pred.\ score & post-hoc predictive score: next-step prediction error of a model trained on generated paths and tested on real ones~\citep{TimeGAN, esteban2017real, PCFGAN}. \\
Sig-MMD & signature-kernel MMD between path distributions~\citep{chevyrev2022signature, PCFGAN}. \\
Sig-$W_{1}$ & signature-based Wasserstein-1 path distance~\cite{HighRank}. \\
ACF & autocorrelation-function distance, per asset and lag. \\
ONND & one-nearest-neighbour distance for distributional fidelity~\cite{HighRank}. \\
Marginal & marginal-distribution histogram distance. \\
Correlation & $L_{1}$ error of cross-asset correlation matrix. \\
Covariance & Frobenius distance of covariance matrices. \\
\midrule
\multicolumn{2}{l}{\emph{Volatility-graph specific}} \\
DY spillover & Diebold-Yilmaz generalised-FEVD volatility-spillover graph~\cite{DieboldYilmaz2012}. \\
\bottomrule
\end{tabularx}
\caption{Metric abbreviations used throughout the task catalogue.}
\label{tab:appendix-metrics}
\end{table}

\subsection{Curated forecasting tasks}\label{app:curated-forecasting}

The forecasting catalogue is shown in Table~\ref{tab:curated-forecasting}. For each task we report the source paper, the data and frequency, the supervised contract (lookback $L$ to horizon $H$), and the headline metric family. Three entries deviate from the standard time-series-tensor and are flagged in the table: Equity excess return uses a tabular per-row input (42 features per asset-month); Global index volatility is graph-conditioned (correlation + Diebold-Yilmaz spillover graphs); and Yield curve is probabilistic (lower / central / upper interval curves).

Following the method of task generation proposed in ~\cite{MOSAIC} and its accompanying code, we construct  38 curated forecasting tasks, which cover U.S.\ equities (organised by sector), U.S.\ Treasury curve points (3m--30y), G7 / emerging-market / Asia-focused FX pairs, crypto majors and altcoins, and metal / energy / soft commodities, and instantiate 7-8 variants per asset family with different panel selections and lookback-to-horison contracts. All variants are scored with the same metric set as in the source paper (RMSE, MAE, and the $\Delta$risk triplet $(\Delta\text{Sharpe},\Delta\text{VaR},\Delta\text{ES})$ between predicted and realised paths).

\begin{table*}[h!]
\centering
\footnotesize
\resizebox{2\columnwidth}{!}{%
\begin{tabular}{l l l l l}
\toprule
\textbf{Task} & \textbf{Source} & \textbf{Data / frequency} & \textbf{Contract} \scriptsize{($L\!\to\!H$)} & \textbf{Metrics} \\
\midrule
LOB                                 & \cite{MOSAIC}                   & 14-channel LOB; high-frequency bars         & $72\!\to\!3$ (5 of 14 price channels scored) & RMSE, MAE, $\Delta$risk \\
Stock-10                            & \cite{MOSAIC}                   & 10 U.S.\ equities; daily                    & $60\!\to\!3$                      & RMSE, MAE, $\Delta$risk \\
Crypto pairs                        & \cite{MOSAIC}                   & 20 crypto pairs; hourly                     & $72\!\to\!3$                      & RMSE, MAE, $\Delta$risk \\
Exchange rate                       & \cite{TSAgent}                  & 8 FX series; daily                          & $60\!\to\!20$                     & RMSE, MAE, MAPE, SMAPE \\
Stock OHLCV                         & \cite{TSAgent}                  & 5-channel OHLCV; daily                      & $60\!\to\!10$ (Close only scored) & RMSE, MAE, MAPE, SMAPE \\
TSFM                                & \cite{TSFM}                     & 60-name U.S.\ excess returns; daily         & $252\!\to\!1$, yearly walk-forward & $R^{2}_{\rm OOS}$, Sharpe, NW $t$-stat \\
NIFTY-50                            & \cite{Nifty50MultiFactor}       & 8-factor panel; daily                       & $30\!\to\!1$                      & $R^{2}$, RMSE, MAPE \\
Equity excess return$^{\ast}$       & \cite{EquityExcessReturnForeca} & U.S.\ stock-month panel, 42 features        & per-row, yearly walk-forward      & $R^{2}_{\rm zero}$, DM, long-short Sharpe \\
Global index volatility$^{\dagger}$ & \cite{GobalIndexVolForeca}      & 8 indices with corr.\ / DY graphs; daily    & $21\!\to\!\{1,5,15,21\}$          & MSE, MAFE, MAPE, $R^{2}$ per horizon \\
Yield curve$^{\ddagger}$            & \citep{YieldCurveForeca}        & Multi-country EIOPA curves; monthly         & $10\!\times\!150\!\to\!150$       & MSE, MAE, PICP, MPIW \\
\midrule
\multicolumn{5}{l}{\emph{MOSAIC pipeline-generated tasks (7--8 variants per asset family)}} \\
Equity panels      & \cite{MOSAIC} & U.S.\ equity panels by sector (5--9 names); daily            & $20\!\to\!1$ -- $90\!\to\!20$  & RMSE, MAE, $\Delta$risk \\
Commodity panels   & \cite{MOSAIC} & metals / energy / softs panels (6--17 names); daily          & $20\!\to\!2$ -- $90\!\to\!20$  & RMSE, MAE, $\Delta$risk \\
Crypto panels      & \cite{MOSAIC} & majors / DeFi / L1 / altcoin panels (6--12 names); hourly    & $12\!\to\!1$ -- $168\!\to\!24$ & RMSE, MAE, $\Delta$risk \\
FX panels          & \cite{MOSAIC} & G7 / EM / Asia / European FX panels (6--10 pairs); daily     & $10\!\to\!1$ -- $90\!\to\!20$  & RMSE, MAE, $\Delta$risk \\
Treasury curve     & \cite{MOSAIC} & U.S.\ Treasury curve segments (3m--30y, 6--10 points); daily & $5\!\to\!1$  -- $60\!\to\!10$  & RMSE, MAE, $\Delta$risk \\
\bottomrule
\end{tabular}%
}
\caption{Forecasting tasks in OpenFinGym. The top block lists the individually curated tasks; the bottom block summarises additional tasks produced by the MOSAIC pipeline, with each row collapsing several variants that share a data family and metric set but differ in panel composition and lookback-to-horizon contract. Input and output channel counts are equal unless noted. $\Delta$risk denotes the triplet $(\Delta\text{Sharpe},\Delta\text{VaR},\Delta\text{ES})$ between predicted and realised paths. Markers: $^{\ast}$~tabular per-row input (no lookback tensor); $^{\dagger}$~graph-conditioned input (correlation and Diebold-Yilmaz spillover graphs alongside the lookback tensor); $^{\ddagger}$~probabilistic output (lower / central / upper interval curves) rather than a point forecast. All metric abbreviations are defined in Table~\ref{tab:appendix-metrics}.}
\label{tab:curated-forecasting}
\end{table*}

\subsection{Curated generative tasks}\label{app:curated-generation}

The generative catalogue is shown in Table~\ref{tab:curated-generation}. The \emph{Sample contract} column pins the shape of the generated tensor: unconditional tasks list a single window shape, while conditional tasks list the context-to-future shape together with the number $K$ of futures generated per context.

\begin{table*}[t]
\centering
\footnotesize
\resizebox{2\columnwidth}{!}{%
\begin{tabular}{l l l l l}
\toprule
\textbf{Task} & \textbf{Source} & \textbf{Data / frequency} & \textbf{Sample contract} & \textbf{Metrics} \\
\midrule
Crypto              & \cite{MOSAIC}    & 20 pairs; hourly             & uncond.\ $60\!\times\!20$              & Marginal, Correlation, ACF, Covariance, $\Delta$risk \\
LOB                 & \cite{MOSAIC}    & 14-channel LOB; HF bars           & uncond.\ $60\!\times\!14$              & Marginal, Correlation, ACF, Covariance  \\
Stock               & \cite{MOSAIC}    & 10 equities; daily           & uncond.\ $60\!\times\!10$              & Marginal, Correlation, ACF, Covariance, $\Delta$risk \\
Crypto log-return   & \cite{TSAgent}   & 10 returns; hourly           & uncond.\ $24\!\times\!10$              & VaR, ES \\
Exchange            & \cite{TSAgent}   & 8 FX series; daily           & uncond.\ $7\!\times\!8$                & Marginal, Correlation, ACF, Covariance \\
Stock log-return    & \cite{TSAgent}   & 8 stocks; daily              & uncond.\ $5\!\times\!8$                & Marginal, Correlation, ACF, Covariance \\
OU                  & \cite{PCFGAN}    & Ornstein--Uhlenbeck          & uncond.\ $64\!\times\!1$               & Disc., Pred., Sig-MMD \\
Rough vol           & \cite{PCFGAN}    & log price + log vol          & uncond.\ $200\!\times\!2$              & Disc., Pred., Sig-MMD, ACF \\
Stock OHLCV         & \cite{PCFGAN}    & 10 stocks OHLCV; daily       & uncond.\ $20\!\times\!5$               & Disc., Pred., Sig-MMD \\
US-5 (cond.)        & \cite{HighRank}  & 5 U.S.\ stocks; daily ret.   & cond.\ $5\!\to\!5$, $K\!=\!128$               & Sig-$W_{1}$, ACF, Correlation, Marginal, Am.-put err., ONND \\
fBM (cond.)         & \cite{HighRank}  & 3D fBM, $H\!=\!0.25$         & cond.\ $6\!\to\!5$, $K\!=\!1$               & Sig-$W_{1}$, ACF, Correlation, Marginal, ONND \\
Tail-GAN            & \cite{TailGAN}   & 5-asset synthetic            & uncond.\ $100\!\times\!5$              & rel.\ VaR / ES, Kupiec, Correlation, ACF \\
TimeGAN stock       & \cite{TimeGAN}   & 6-column stock; daily      & uncond.\ $24\!\times\!6$               & Disc., Pred. \\
\bottomrule
\end{tabular}%
}
\caption{Curated generative tasks. The \emph{Sample contract} column reports the window shape as time-steps $\times$ channels for unconditional tasks; conditional bundles list context length $\to$ future length together with the number $K$ of generated futures emitted per context. Metric symbols are defined in Table~\ref{tab:appendix-metrics}; $\Delta$risk denotes the triplet $(\Delta\text{Sharpe},\Delta\text{VaR},\Delta\text{ES})$.} For the fBM task, the model emits a single future per context ($K\!=\!1$). For US-5, the American-put error aggregates over futures, therefore uses K = 128; the put is priced on the $K\!=\!128$ paths at an at-the-money strike.
\label{tab:curated-generation}
\end{table*}

\subsection{Curated Fraud detection tasks}
\label{app:fraud-detecction}
The fraud detection tasks are shown in Table~\ref{tab:fraud_detect}. It contains seven graph-structured tasks drawn from established anomaly-detection and fraud-detection benchmarks. Six tasks are node-level anomaly scoring problems on attributed graphs; one task is a graph-level binary classification on a synthetic motif benchmark.

\begin{table*}[h]
    \centering
    \resizebox{2\columnwidth}{!}{%
    \begin{tabular}{l l p{5.5cm} p{2.2cm} l p{2.8cm}}
        \toprule
        \textbf{Dataset} & \textbf{Original task} & \textbf{Description} & \textbf{Evaluation mode} & \textbf{Anomaly rate} & \textbf{Metric} \\
        \hline
        Amazon & \cite{zheng2025cluster} & 10,224-node e-commerce product graph; 25 feat.; COO adjacency & Node anomaly scoring & 6.8\% & AUCROC, Recall \\
        BlogCatalog & \cite{zheng2025cluster} & 5,196-node social network (bloggers); 8,189 feat.; COO adjacency & Node anomaly scoring & 5.8\% & AUCROC, Recall \\
        \hline
        YelpChi & \cite{zheng2025cluster} & 23,831-node review graph (reviews); 32 feat.; COO adjacency & Node anomaly scoring & 5.1\% & AUCROC, Recall \\
        \hline
        DBLP & \cite{zheng2025cluster} & 4,057-node citation/co-author graph; 334 feat.; 2 relation types merged into dense adjacency & Node anomaly scoring & 7.0\% & AUCROC, Recall \\
        \hline
        IMDB 5K & \cite{zheng2025cluster} & 4,780-node movie co-occurrence graph; 1,232 feat.; 2 relation types merged into dense adjacency & Node anomaly scoring & 7.0\% & AUCROC, Recall \\
        \hline
        CERT & \cite{zheng2025cluster} & 1,000-node user behaviour graph; 6 feat.\ (2 per view: email, logon, file); COO email-activity adjacency & Node anomaly scoring & 7.0\% & AUCROC, Recall \\
        \hline
        BA-2motifs & \cite{li2024sefraud} & 1,000 synthetic graphs; 25 nodes $\times$ 10 feat.\ per graph (constant 0.1); 25 $\times$ 25 adjacency per graph; balanced binary classes (house vs.\ 5-cycle motif) & Graph binary classification & 50.0\% (balanced) & AUCROC, Recall \\
        \hline
    \end{tabular}
    }
    \caption{Fraud detection tasks in OpenFinGym. The Anomaly rate column reports the fraction of positive (anomalous / fraud-class) labels in the training split. }
    \label{tab:fraud_detect}
\end{table*}

\end{document}